\pdfoutput=1
\documentclass[11pt]{article}

\usepackage[final]{acl}

\usepackage{times}
\usepackage{latexsym}
\usepackage{breakcites}
\usepackage[T1]{fontenc}

\usepackage[utf8]{inputenc}

\usepackage{microtype}

\usepackage{inconsolata}

\usepackage{graphicx}
\usepackage{booktabs}
\usepackage{tabularx}
\usepackage{multirow}
\usepackage{scalerel}
\usepackage{float}

\title{\textbf{MiRAGeNews:} \textbf{M}ult\textbf{i}modal \textbf{R}ealistic \textbf{A}I-\textbf{Ge}nerated \textbf{News} Detection}

\author{Runsheng  (Anson)  Huang, \quad
  Liam Dugan, \quad
  Yue Yang, \quad
  Chris Callison-Burch \\
  University of Pennsylvania \\
  {\tt \{rhuang99,ldugan,yueyang1,ccb\}@seas.upenn.edu}
}

\begin{document}
\maketitle
\begin{abstract}
The proliferation of inflammatory or misleading ``fake'' news content has become increasingly common in recent years. Simultaneously, it has become easier than ever to use AI tools to generate photorealistic images depicting any scene imaginable. Combining these two---AI-generated fake news content---is particularly potent and dangerous. To combat the spread of AI-generated fake news, we propose the MiRAGeNews Dataset, a dataset of 12,500 high-quality real and AI-generated image-caption pairs from state-of-the-art generators. We find that our dataset poses a significant challenge to humans (60\% F-1) and state-of-the-art multimodal LLMs (< 24\% F-1). Using our dataset, we train a multi-modal detector (MiRAGe) that improves by +5.1\% F-1 over state-of-the-art baselines on image-caption pairs from out-of-domain image generators and news publishers. We release our code and data to aid future work on detecting AI-generated content.\footnote{\url{https://github.com/nosna/miragenews}}
\end{abstract}

\section{Introduction}

Diffusion models \cite{ho2020denoising} have shown remarkable advancements in generating hyper-realistic images. Particularly, models like \href{https://www.midjourney.com}{Midjourney} can produce images that even graduate CS students cannot distinguish (Sec \ref{sec:humaneval}). This capability has profound implications, especially with regard to the dissemination of disinformation. Recently, there has been a noticeable surge in AI-generated news images on social media \cite{metz_i_2024}. When coupled with the proficiency of large language models (LLMs) in generating grammatically and contextually appropriate captions, the potential for AI-driven disinformation campaigns becomes an increasingly critical concern.

Recent work on detecting AI-generated images has shown impressive performance on images generated by models such as Stable Diffusion \cite{sd}, Glide \cite{glide}, and DALL-E 2 \cite{dalle2}. However, these images are not always realistic and are often easily distinguishable by humans due to their obvious anomalies. %(See Appendix for comparison). 
Since the datasets used in previous works do not accurately represent the current challenge posed by state-of-the-art (SOTA) diffusion-based models, there is an evident need for a challenging dataset comprising realistic AI-generated news images and captions that provide the research community with a foundation to develop and test new detection methods.

\begin{figure}
    \centering
    \includegraphics[width=\columnwidth]{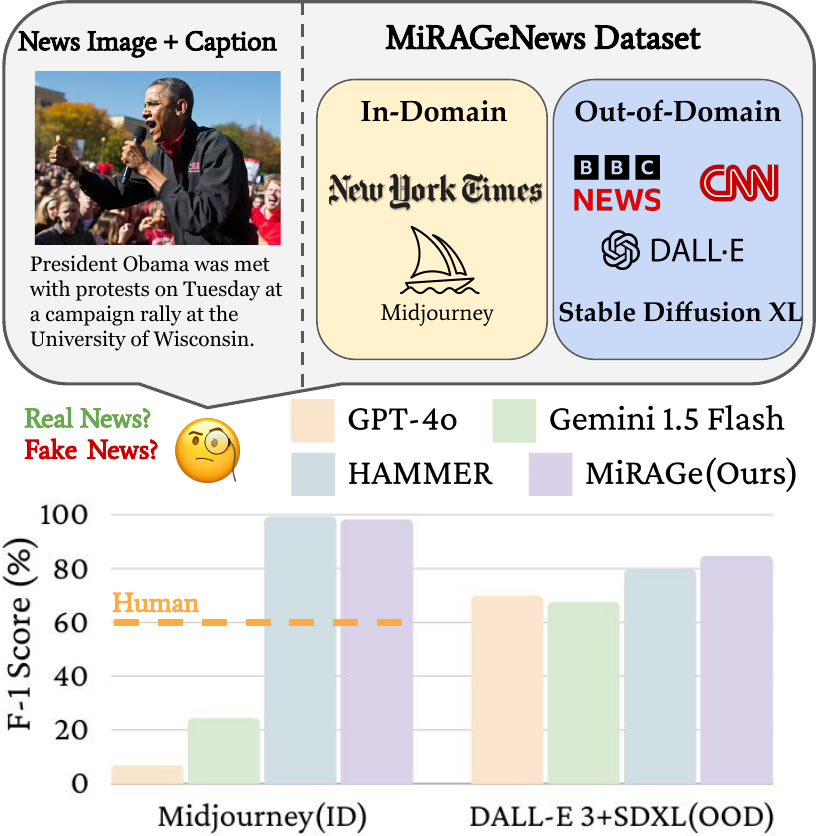}
    \vspace{-.6cm}
    \caption{Multimodal fake news with hyperrealistic generated images from Midjourney poses a significant challenge for both state-of-the-art MLLMs (< 24\% F-1) and humans (60\% F-1). Our detectors achieve over 98\% F-1 on in-domain (ID) data and can generalize on out-of-domain (OOD) data from unseen news publishers and image generators (85\% F-1)}
    \label{fig:fig1}
    \vspace{-.3cm}
\end{figure}

In this work, we present the \textbf{MiRAGeNews Dataset}, a dataset of real and fully generated news image-caption pairs with 12,500 generated images and corresponding captions from SOTA generators as training and validation sets. To evaluate detectors' out-of-domain robustness, we also create a test set of 2,500 image-caption pairs from various unseen image generators and news publishers. 

Using this data, we train \textbf{MiRAGe}, a multimodal detector that fuses an image detector and a text detector, both of which are ensembles of a black-box linear model and an interpretable concept bottleneck model. We show that MiRAGe exhibits better out-of-distribution (OOD) robustness compared to previous state-of-the-art detectors and MLLMs.

\section{MiRAGeNews Dataset}

\subsection{Dataset Creation}

\paragraph{Real Images and Captions.} To create our data, we first sample 6,500 New York Times image and caption pairs from TARA \cite{tara} as our ``real'' news subset. We select TARA for this due to the presence of specific information on the location and time of the news events in the captions. This geographical and temporal information helps provide extra constraints to the model during generation.

\paragraph{Fake Caption Generation.} To simulate instances of real-world disinformation, we explicitly prompt GPT-4 \cite{openai2024gpt4} to take a real caption and "\textit{generate fictional captions that could be considered harmful or misleading}". We also ask it to incorporate all named entities from the original caption to ensure the generated caption does not stray too far from the original.

\paragraph{Fake Image Generation.} We choose Midjourney V5.2 as the image generator, considering its hyper-realistic generations and relatively lenient moderation.\footnote{Other image generators have stricter rules with regard to the generation of harmful imagery (i.e., fabricated events of public figures or graphic crime scenes).} To generate fake images, we use fake captions from GPT-4 as prompts with additional keywords to restrict the photo style. We also include the aspect ratio of the corresponding real images in the prompt to reflect the realistic property of news images.\footnote{Midjourney prompt: "\{\textit{caption}\} News photo style --ar width:height --style raw".}

\begin{figure}
    \centering
    \includegraphics[width=\columnwidth]{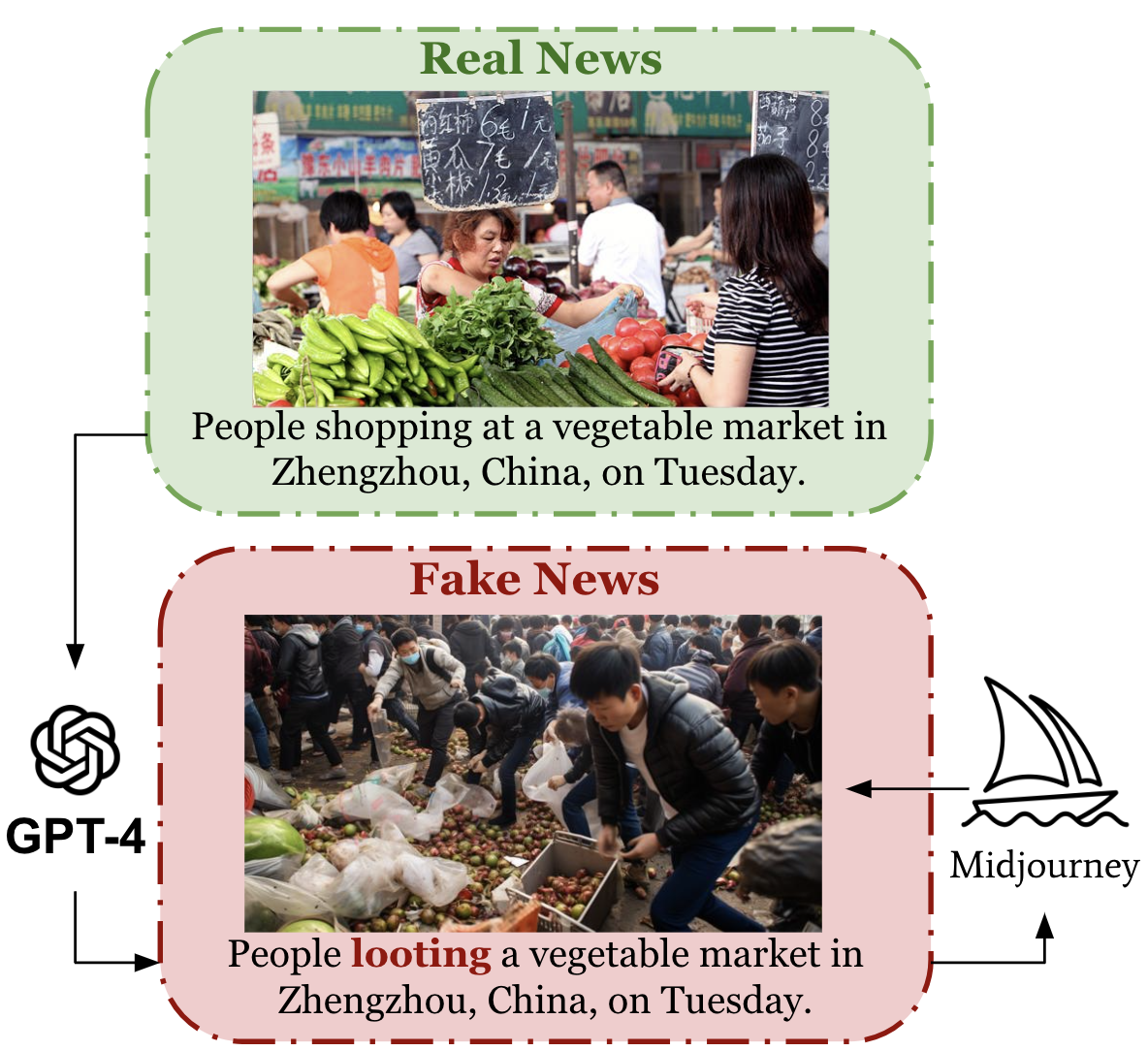}
    \caption{Example of MiRAGeNews dataset generation. We use GPT-4 to generate a misleading caption, which is then used by Midjourney to generate the image.}
    \label{fig:datagen}
\end{figure}

\subsection{Task Design}
Our detection task is designed to simulate the real-world scenario where news on social media is often presented as an image-caption pair. The detector needs to determine if both the image and caption are real (label 0) or if both are generated (label 1). 

To evaluate the detector's generalization ability, we also collect 250 image-caption pairs each from BBC and CNN\footnote{\href{https://www.kaggle.com/datasets/szymonjanowski/internet-articles-data-with-users-engagement}{https://www.kaggle.com/datasets/szymonjanowski/internet-articles-data-with-users-engagement}} to simulate the domain gaps of news content from different news publishers. We follow the same process to generate fake captions and use unseen generative models, DALL-E 3 and Stable Diffusion XL (SDXL), to generate Out-of-Domain (OOD) fake images. We apply every combination of unseen news and image generators to obtain four OOD datasets: BBC + DALL-E 3, CNN + DALL-E 3, BBC + SDXL, and CNN + SDXL. With the addition of 500 in-domain real or fake image-caption pairs, we construct our test set with 5 small datasets, totaling 2,500 image pairs.

\begin{figure*}[t]
    \centering
    \includegraphics[width=\textwidth]{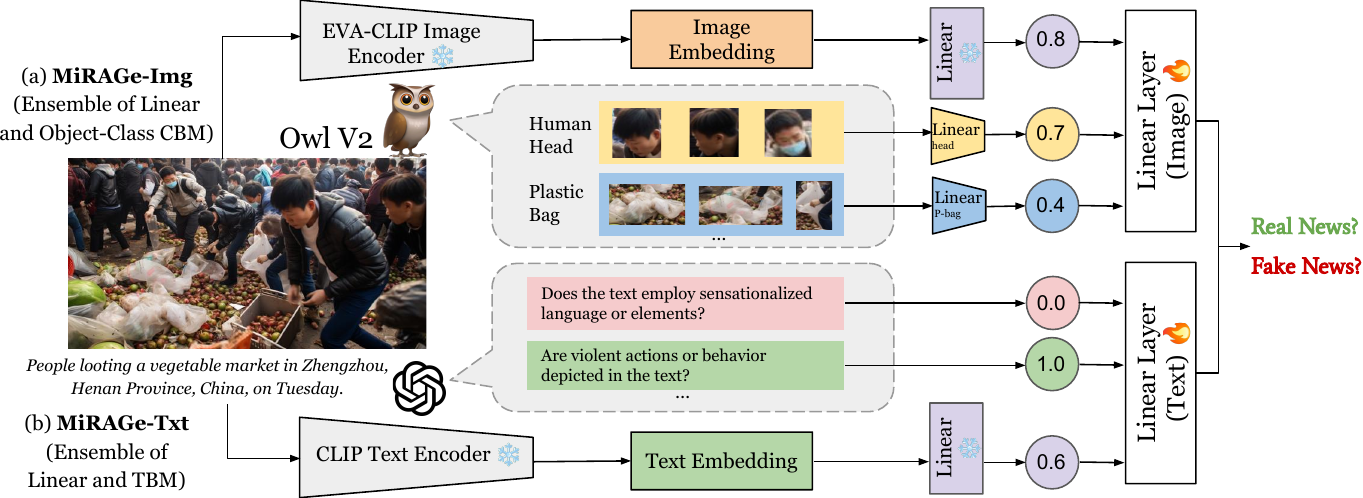}
    \caption{Overview of our \textbf{MiRAGe} detector for multimodal AI-generated news detection, which combines \textbf{MiRAGe-Img} with \textbf{MiRAGe-Txt}. \textbf{MiRAGe-Img} trains a linear layer on the outputs from the image linear model and Object-Class Concept Bottleneck Model (CBM), while \textbf{MiRAGe-Txt} trains a linear layer on the outputs from the text linear model and Text Bottleneck Model (TBM). Outputs from two models can be either early fused or late fused to make the final prediction on the image-caption pair.}
    \label{fig:ensembles}
\end{figure*}

\subsection{Human Evaluation}
\label{sec:humaneval}

 To evaluate human detection capability on our dataset, we recruited 112 students who are taking a graduate-level NLP course with extra credit as compensation. Each student is randomly assigned 20 image caption pairs and is asked to separately determine if the image is generated and if the caption is generated. Each pair in our survey was given to three participants, and we used a majority vote to determine the final prediction by humans.

Our evaluation results aligned with our hypothesis that humans are not good at this detection task: they detected only 60.3\% of the generated images and 53.5\% of the generated captions. The well-educated participants are representative of a high-performing subpopulation yet performing approximately equally with random guessing, which implies that these realistic fake news stories are fully capable of fooling humans, and we need models that can assist in this task to combat disinformation.

\section{Detectors}

\paragraph{Baselines.} We compare our detectors against recent baselines in three different settings: Image-only, Text-only, and Multimodal detection. For image-only, we compare to DE-FAKE \cite{defake}, DIRE \cite{wang2023dire}, and KNN \cite{ojha23}. For text-only, we compare with the Text Bottleneck Model (TBM) \cite{tbm}, and for Multimodal, we compare with HAMMER \cite{dgm4}. We also test simple linear models in each modality and benchmark state-of-the-art MLLMs on the image-only and multimodal detection tasks. See Appendix \ref{sec:appendix} for a more detailed discussion on each of the detectors tested.

\paragraph{MiRAGe-Img} trains a linear layer on top of the outputs from two models: (1) a linear model trained using EVA-CLIP \cite{evaclip} image embeddings and (2) an Object-Class Concept Bottleneck Model (CBM) containing 300 object-class classifiers trained on crops of different objects from Owl V2 \cite{owlv2}. The interpretable Object-Class CBM focuses on regional anomalies, while the linear model captures the global features.

\paragraph{MiRAGe-Txt} trains a linear layer on top of the outputs from two models: (1) a linear model that is trained using CLIP text embeddings \cite{clip} and (2) a Text Bottleneck Model \cite{tbm} that extracts 18 textual concepts in the caption. Similar to MiRAGe-Img, we incorporate the interpretable concept-based approach to capture the auxiliary signals that the linear models missed.

\begin{figure}[t]
    \centering
    \includegraphics[width=\columnwidth]{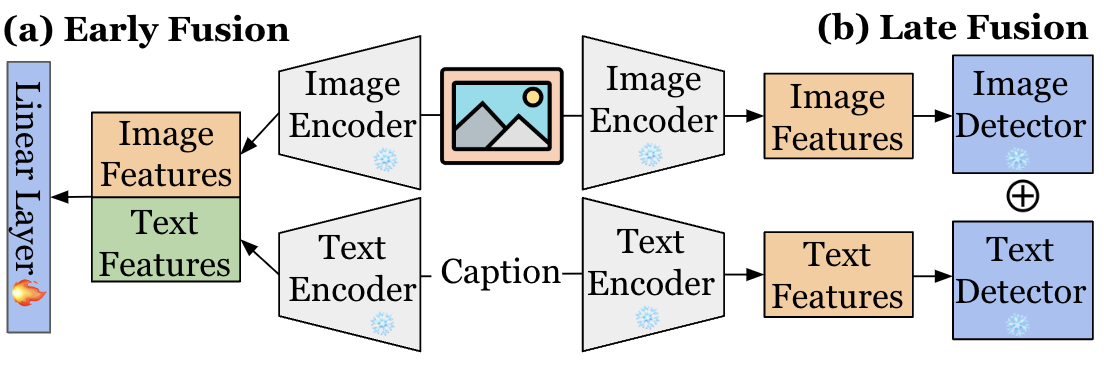}
    \caption{\textbf{(a) Early Fusion} detector uses both image and text features together for classification while \textbf{(b) Late Fusion} detector uses outputs from previously trained unimodal detectors.}
    \label{fig:fus}
\end{figure}

\paragraph{MiRAGe} fuses MiRAGe-Img and MiRAGe-Txt together for multimodal generated news detection.
We apply two fusion techniques as illustrated in Figure \ref{fig:fus}: \textbf{Early Fusion} involves concatenating the outputs from MiRAGe-Img and MiRAGe-Txt before training a linear layer for classification, while \textbf{Late Fusion} computes a final prediction from the outputs of these two models with no extra training.

We train all models until the evaluation loss plateaus and apply the classification threshold that gives the highest evaluation accuracy in testing. As for our design decisions, we conducted a detailed ablation study for each part of the MiRAGe detector in Appendix \ref{sec:ablation}. Due to the results of these ablations, we chose the late fusion model as our final \textbf{MiRAGe} detector.

\section{Results}

\begin{figure}
    \centering
    \includegraphics[width=\columnwidth]{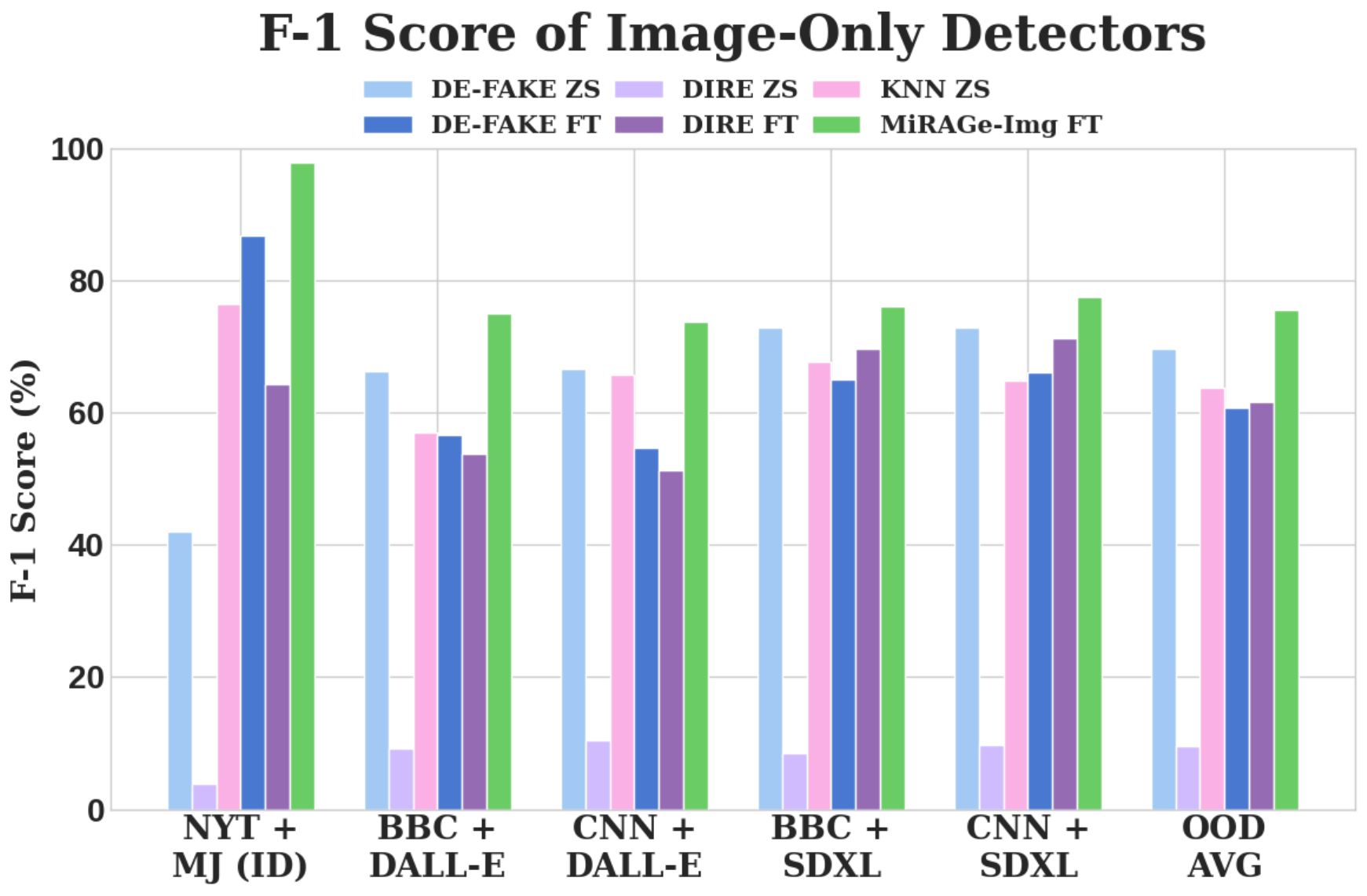}
    \caption{We see that MiRAGe-Img outperforms existing image-only detectors in both in-domain (ID) and out-of-domain (OOD). ZS and FT are short for Zero-Shot and Fine-Tuned, respectively}
    \label{fig:bars}
\end{figure}

\subsection{Image-only}
As shown in Figure \ref{fig:bars}, the MiRAGe-Img model demonstrates better in-domain (ID) performance and out-of-domain (OOD) generalization ability than our baselines. 
We find that zero-shot DIRE struggles on all datasets, most likely due to the major domain shift from training data (bedroom images) to testing data (news images). While the models fine-tuned on ID data have substantially lower performance on DALL-E, we are surprised to find that DIRE FT has a higher average F-1 on SDXL (70.5\%) than Midjourney (64.4\%). One reasonable speculation could be a similar base model shared by two generators, which is reflected in similar reconstruction errors. Our model shows that using the Object-Class CBM along with the linear model helps improve OOD robustness.

\subsection{Text-only}
As shown in Table \ref{tab:example}, MiRAGe-Txt outperforms baselines in both ID captions rewritten from New York Times news and OOD captions from unseen news publishers (BBC and CNN).

\subsection{Multimodal}
\begin{figure}
    \centering
    \includegraphics[width=\columnwidth]{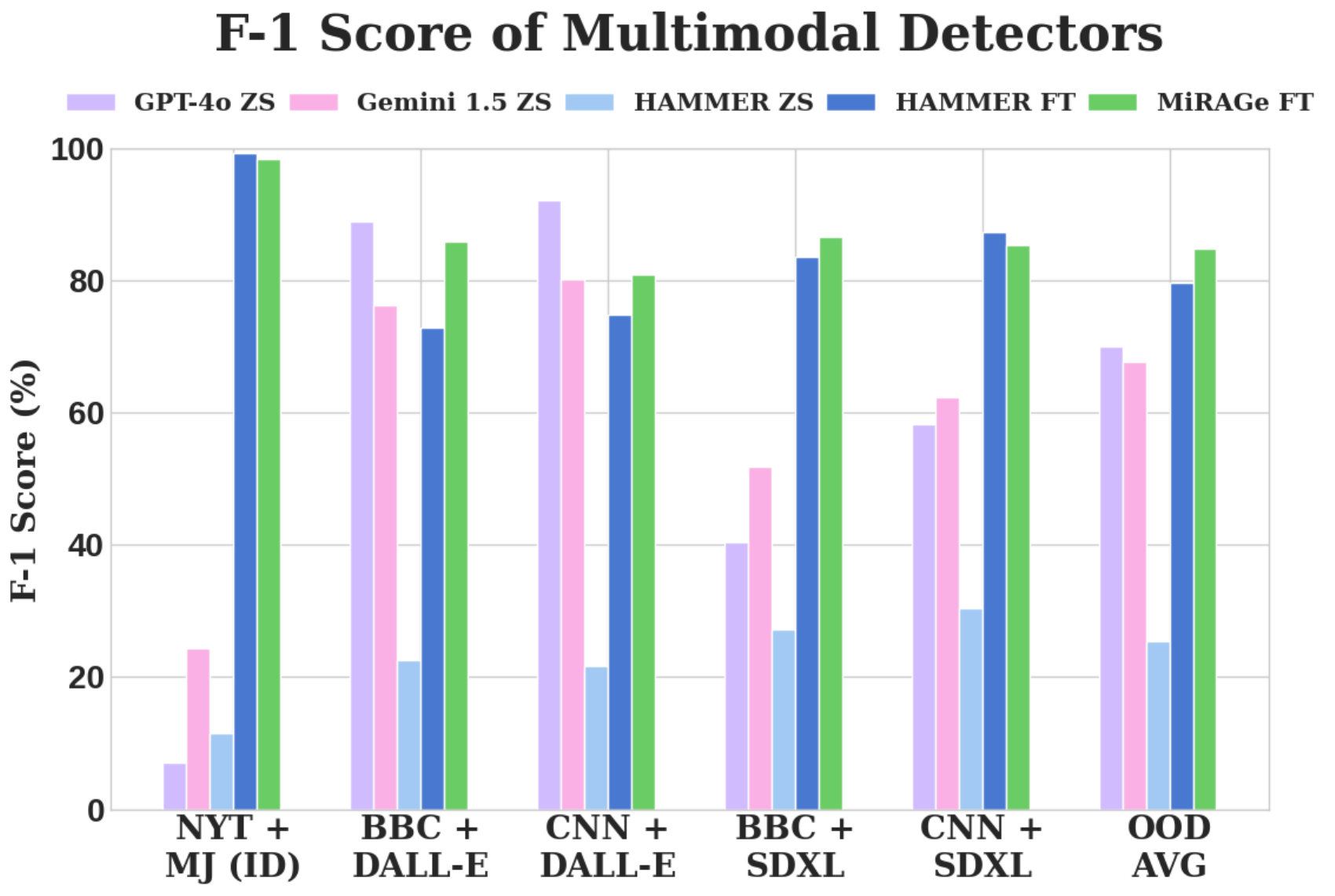}
    \caption{We see that MiRAGe outperforms existing image-text detectors in out-of-domain settings. ZS and FT are short for Zero-Shot and Fine-Tuned, respectively.}
    \label{fig:bars2}
\end{figure}

In the multimodal setting with both images and captions, our MiRAGe detector exhibits better OOD robustness than our baselines (see Figure \ref{fig:bars2}). Both state-of-the-art MLLMs (GPT-4o and Gemini 1.5) struggle on ID data. We further find that the low F1 attribute to extremely low recall (fake accuracy) as shown in Table \ref{tab:accuracy}. However, GPT-4o's surprising performance on DALL-E makes us speculate it might be trained with DALL-E images and that MLLMs' zero-shot performance heavily varies depending on the training data distribution. While utilizing additional signals from semantic inconsistency between images and texts helps HAMMER generalize on OOD data, MiRAGe, which fuses MiRAGe-Img and MiRAGe-Txt, has shown better OOD performance overall (see Appendix \ref{sec:ablation}).

\section{Related Work}

\paragraph{Multimodal Fake News Datasets.} There are many datasets for detecting generated images \cite{wang2020cnngenerated, he2021forgerynetversatilebenchmarkcomprehensive, zhu2023genimagemillionscalebenchmarkdetecting} and generated text \cite{dugan2024raidsharedbenchmarkrobust, li2024magemachinegeneratedtextdetection, wang2024m4multigeneratormultidomainmultilingual}. However, there are relatively few publicly available datasets for detecting generated image-text pairs---especially in the news domain. The Twitter MediaEval dataset \cite{mediaeval} contains a corpus of 17,000 tweets on two events with 514 real or fake images. Weibo \cite{weibo} collects real and rumor news posts that are verified by the authoritative debunking system with mostly real images. FakeNewsNet \cite{fakenewsnet} collects real and fake news from Politifact and GossipCop and contains 17,214 human-written news articles with images and 1,986 fake news articles with images. The more recent DGM4 dataset \cite{shao2023detectinggroundingmultimodalmedia} offers 230k news samples with image and text, which contain 77k pristine pairs and 152k manipulated pairs from small regional manipulations on image and/or text. While previous datasets offer the foundation of multi-modal fake news, our dataset aims to address the more dangerous forms of fake news, namely the ones that have convincingly deceptive visuals fully generated by diffusion models.

\paragraph{Generated Image Detection.} Many previous works have explored different methods for effective fake news detection \cite{eann, mvae, spotfake}. 
Stemming from detecting GAN-generated images \cite{gragnaniello_are_2021,wang2020cnngenerated}, 
many methods have been proposed to detect diffusion-based generation. DE-FAKE \cite{defake} uses BLIP to generate a caption for every image, then combines both features to predict. DIRE \cite{wang2023dire} analyzes the reconstruction error during denoising. \citet{ojha23} computes the distance between the testing image and the training set and uses KNN to predict real vs. fake. Compared to these black-box methods, our proposed object-class CBM offers a new perspective on interpretable generated image detectors. Moreover, with the addition of generated fake captions, our dataset lays the groundwork for more creative future works.

\section{Conclusion}

In this paper, we introduce MiRAGeNews, a dataset designed to facilitate the development and benchmarking of detection methods for AI-generated news. Our dataset is the first of its kind to include high-quality images from modern generators along with misleading or harmful captions, and our results highlight the significant challenges faced by humans and current state-of-the-art multimodal language models in detecting such news content. We show that classifiers trained on our data achieve high accuracy on the most difficult-to-detect images while still showing strong generalization performance on out-of-domain generators and news sources.

\section{Limitations}
In the design of the testing set with OOD data, while both our real images and fake images are OOD, it is not truly OOD for captions. We use the same procedure to prompt GPT-4 when generating, and the domain shift, if any, will come from the real captions of unseen news publishers.
Our experiments would be more comprehensive if we finetune GPT-4o and Gemini 1.5 on our dataset. However, training with images that contain public figures and faces would violate the Term of Service, thus making our dataset mostly unavailable.
Also, since our real news dataset is mostly from the New York Times, all of our real and fake captions are English only, making it a monolingual dataset. Although it is possible to machine translate the entire dataset, we would leave this to future work.

\section{Ethics Statement}
Since most of the fake images from our dataset are generated from misleading or harmful captions, and Midjourney's moderation system is not perfect, some generations might be considered to be unsafe. Although the real captions that GPT -4 used during the generation are dated, it is still very likely that the generated caption can stand alone as a source of disinformation about current events.

\section*{Acknowledgement}
We thank the students from CIS 5300 at the University of Pennsylvania for the human annotation on generated image-caption pairs. We also thank Muzi (Hex) Li for her support on the project. This research is supported in part by the Office of the Director of National Intelligence (ODNI), Intelligence Advanced Research Projects Activity (IARPA), via the HIATUS Program contract \#2022-22072200005. The views and conclusions contained herein are those of the authors and should not be interpreted as necessarily representing the official policies, either expressed or implied, of ODNI, IARPA, or the U.S. Government. The U.S. Government is authorized to reproduce and distribute reprints for governmental purposes, notwithstanding any copyright annotation therein.

% Bibliography entries for the entire Anthology, followed by custom entries
%\bibliography{anthology,custom}
% Custom bibliography entries only
\bibliography{acl_latex}

\clearpage

\begin{table*}[t]
    \small
    \centering
    \setlength{\tabcolsep}{2.3pt} % Adjusts the space between columns
    \begin{tabular}{c p{1.7cm} c|ccc|ccc|ccc|ccc|ccc|ccc}
    \toprule 
    & & &\multicolumn{3}{c|}{\textbf{NYT+MJ (ID)}}&\multicolumn{3}{c|}{\textbf{BBC+DALL-E}}&\multicolumn{3}{c|}{\textbf{CNN+DALL-E}}&\multicolumn{3}{c|}{\textbf{BBC+SDXL}}&\multicolumn{3}{c|}{\textbf{CNN+SDXL}}&\multicolumn{3}{c}{\textbf{OOD-AVG}}\\
    & & &\textbf{Acc}&\textbf{F1}&\textbf{AP}&\textbf{Acc}&\textbf{F1}&\textbf{AP}&\textbf{Acc}&\textbf{F1}&\textbf{AP}&\textbf{Acc}&\textbf{F1}&\textbf{AP}&\textbf{Acc}&\textbf{F1}&\textbf{AP}&\textbf{Acc}&\textbf{F1}&\textbf{AP}\\
    \midrule
    \multirow{8}{*}{\rotatebox{90}{\textbf{Image}}} & 
    GPT-4o & ZS & 51.6 & 6.2 & 100 & \textbf{93.0} & \textbf{92.7} & 97.4 & \textbf{96.2} & \textbf{96.2} & 96.4 & 67.2 & 52.9 & 93.9 & \textbf{78.2} & 73.5 & 93.8 & 83.7 & 78.8 & 95.4 \\
    & Gemini 1.5 & ZS & 48.0 & 12.2 & 39.1 & 75.9 & 75.0 & 91.7 & 83.8 & 85.6 & 93.4 & 47.7 & 22.3 & 67.6 & 45.4 & 30.1 & 76.6 & 63.2 & 53.3 & 82.3 \\
    & \multirow{2}{*}{DE-FAKE} & \multicolumn{1}{|c|}{ZS} & 61.4 & 42.0 & 35.6 & 63.8 & 66.4 & 58.8 & 61.6 & 66.6 & 50.5 & 69.4 & 73.0 & 60.1 & 67.4 & 73.0 & 63.4 & 65.6 & 69.8 & 58.2 \\
    & & \multicolumn{1}{|c|}{FT} & 87.6 & 86.8 & 33.7 & 44.6 & 56.7 & 47.7 & 44.6 & 54.7 & 40.2 & 52.6 & 65.1 & 63.8 & 55.0 & 66.1 & 56.6 & 49.2 & 60.7 & 52.1 \\
    \addlinespace[0.1ex]
    & \multirow{2}{*}{DIRE} & \multicolumn{1}{|c|}{ZS} & 50.0 & 3.8 & 54.0 & 48.6 & 9.2 & 52.1 & 48.2 & 10.4 & 48.9 & 48.4 & 8.5 & 48.7 & 48.0 & 9.7 & 49.0 & 48.3 & 9.5 & 49.7 \\
    & & \multicolumn{1}{|c|}{FT} & 57.8 & 64.4 & 56.0 & 48.8 & 53.8 & 56.9 & 48.0 & 51.3 & 50.7 & 62.4 & 69.8 & 73.2 & 64.6 & 71.3 & 68.3 & 56.0 & 61.6 & 62.3 \\
    & KNN & ZS & 73.8 & 76.5 & 69.4 & 43.2 & 57.1 & 45.9 & 53.8 & 65.8 & 52.2 & 57.2 & 67.8 & 54.3 & 51.0 & 64.9 & 50.6 & 51.3 & 63.9 & 50.8 \\
    & \textbf{MiRAGe-I} & FT & \textbf{98.0} & \textbf{98.0} & \textbf{99.9} & 77.6 & 75.0 & \textbf{88.4} & 74.6 & 73.8 & \textbf{84.4} & \textbf{78.4} & \textbf{76.1} & \textbf{87.7} & 77.6 & \textbf{77.6} & \textbf{83.0} & \textbf{77.1} & \textbf{75.6} & \textbf{85.9} \\
    \midrule
    % & Linear & FT & 97.6 & 97.6 & \textbf{99.9} & 70.4 & 63.0 & 84.2 & 66.6 & 62.3 & 75.1 & 77.2 & 73.7 & 87.3 & 74.2 & 73.2 & 81.2 & 72.1 & 68.1 & 82.0 \\
    % & CBM & FT & 94.2 & 94.2 & 98.1 & 66.6 & 70.3 & 82.8 & 72.0 & 74.4 & \textbf{84.6} & 65.6 & 69.2 & 74.9 & 65.0 & 65.8 & 73.8 & 67.3 & 69.9 & 79.0 \\

    \multirow{4}{*}{\rotatebox{90}{\textbf{Text}}} & Linear (SB) & FT & 81.8 & 81.8 & \textbf{91.0} & 68.8 & 75.3 & \textbf{85.2} & 70.0 & 75.9 & 87.1 & & & & & & & 69.4 & 75.6 & 86.2 \\
    & Linear (CL) & FT & 81.8 & 80.5 & 90.2 & 68.4 & 74.8 & 84.4 & 77.6 & 80.3 & 90.9 & & & & & & & 73.0 & 77.6 & 87.7 \\
    & TBM & FT & 74.0 & 71.4 & 78.1 & 68.8 & 68.7 & 72.1 & 76.2 & 76.8 & 75.2 & & & & & & & 72.5 & 72.8 & 73.7 \\
    % & TBM+SB & FT & \textbf{83.4} & \textbf{82.4} & \textbf{91.0} & \textbf{73.2} & \textbf{77.1} & \textbf{85.4} & 74.8 & 78.2 & 88.3 & & & & & & & 74.0 & 77.7 & 86.9 \\
    & \textbf{MiRAGe-T} & FT & \textbf{83.2} & \textbf{81.9} & \textbf{91.0} & \textbf{72.6} & \textbf{77.0} & \textbf{85.2} & \textbf{78.4} & \textbf{80.9} & \textbf{92.0} & & & & & & & \textbf{75.5} & \textbf{79.0} & \textbf{88.6} \\
    \midrule
    
    \multirow{5}{*}{\rotatebox{90}{\textbf{Image+Text}}} & GPT-4o & ZS & 51.8 & 7.0 & 100 & \textbf{90.1} & \textbf{89.0} & 100 & \textbf{92.8} & \textbf{92.2} & 100 & 62.7 & 40.4 & 100 & 70.7 & 58.2 & 100 & 79.1 & 70.0 & 100 \\
    & Gemini 1.5 & ZS & 56.4 & 24.4 & 94.6 & 78.9 & 76.4 & 87.2 & 81.2 & 80.2 & 85.1 & 64.2 & 51.8 & 79.2 & 69.1 & 62.4 & 79.4 & 73.4 & 67.7 & 82.7 \\
    & \multirow{2}{*}{HAMMER} & \multicolumn{1}{|c|}{ZS} & 39.0 & 11.6 & 37.2 & 49.2 & 22.6 & 48.7 & 45.2 & 21.7 & 42.9 & 51.0 & 27.3 & 51.4 & 48.8 & 30.4 & 47.5 & 48.6 & 25.5 & 47.6 \\
    & & \multicolumn{1}{|c|}{FT} & \textbf{99.4} & \textbf{99.4} & \textbf{100} & 77.2 & 73.0 & 86.2 & 78.8 & 74.8 & 89.4 & 85.0 & 83.7 & 93.4 & \textbf{88.2} & \textbf{87.4} & \textbf{96.2} & 82.3 & 79.7 & 91.3 \\

    & \textbf{MiRAGe} & FT & 98.4 & 98.4 & 99.9 & 85.4 & 86.0 & \textbf{94.1} & 79.6 & 81.0 & \textbf{90.4} & \textbf{86.0} & \textbf{86.7} & \textbf{95.9} & 83.8 & 85.5 & 94.7 & \textbf{83.7} & \textbf{84.8} & \textbf{93.8} \\
    
    \bottomrule
    \end{tabular}
    \caption{\textbf{The Accuracy, F-1 score, and Average Precision (AP) of all detectors across MiRAGeNews test sets.} Note that AP for GPT-4o and Gemini 1.5 are just precision and not comparable to other models. ZS, FT, SB, and CL stand for zero-shot, fine-tuned, SentenceBert, and CLIP, respectively. We leave SDXL columns blank in Text-only detectors since they share the same textual data as the DALL-E columns. We see that zero-shot models struggle on in-domain data, while fine-tuned models achieve strong in-domain performance and respectable generalization on OOD sections. See Figure \ref{fig:bars} and \ref{fig:bars2} for graphical highlights from this table.}
    \label{tab:example}
\end{table*}

\begin{table*}[ht]
    \small
    \centering
    \setlength{\tabcolsep}{3pt} % Adjusts the space between columns
    \begin{tabular}{c p{1.7cm} c|cc|cc|cc|cc|cc|cc}
    \toprule 
    & & &\multicolumn{2}{c|}{\textbf{NYT+MJ (ID)}}&\multicolumn{2}{c|}{\textbf{BBC+DALL-E}}&\multicolumn{2}{c|}{\textbf{CNN+DALL-E}}&\multicolumn{2}{c|}{\textbf{BBC+SDXL}}&\multicolumn{2}{c|}{\textbf{CNN+SDXL}}&\multicolumn{2}{c}{\textbf{OOD-AVG}}\\
    & & &\textbf{Real}&\textbf{Fake}&\textbf{Real}&\textbf{Fake}&\textbf{Real}&\textbf{Fake}&\textbf{Real}&\textbf{Fake}&\textbf{Real}&\textbf{Fake}&\textbf{Real}&\textbf{Fake}\\
    \midrule
    \multirow{10}{*}{\rotatebox{90}{\textbf{Image}}} 
    & GPT-4o & ZS & \textbf{100.0} & 3.2 & \textbf{97.6} & \textbf{88.4} & \textbf{96.4} & \textbf{96.0} & \textbf{97.6} & 36.8 & \textbf{96.0} & 60.4 & \textbf{96.9} & 70.4 \\
    & Gemini 1.5 & ZS & 88.8 & 7.2 & 88.4 & 63.4 & 88.6 & 79.0 & 82.0 & 13.4 & 72.0 & 18.8 & 82.8 & 43.7 \\
    & \multirow{2}{*}{De-Fake} & ZS & 94.8 & 28.0 & 56.0 & 71.6 & 46.8 & 76.4 & 56.0 & 82.8 & 46.8 & 88.0 & 51.4 & 79.7 \\
    &  & FT & 94.0 & 81.2 & 16.8 & 72.4 & 22.4 & 66.8 & 16.8 & 88.4 & 22.4 & 87.6 & 19.6 & 78.8 \\
    \addlinespace[0.1ex]
    & \multirow{2}{*}{DIRE} & ZS & 98.0 & 2.0 & 92.0 & 5.2 & 90.4 & 6.0 & 92.0 & 4.8 & 90.4 & 5.6 & 91.2 & 5.4 \\
    &  & FT & 39.2 & 76.4 & 38.0 & 59.6 & 41.2 & 54.8 & 38.0 & 86.8 & 41.2 & 88.0 & 39.6 & 72.3 \\
    & KNN & ZS & 62.4 & 85.2 & 10.8 & 75.6 & 18.8 & 88.8 & 24.4 & \textbf{90.0} & 11.2 & \textbf{90.8} & 16.3 & \textbf{86.3} \\

    & \textbf{MiRAGe-I} & FT & 98.8 & \textbf{97.2} & 88.0 & 67.2 & 77.6 & 71.6 & 88.0 & 68.8 & 77.6 & 77.6 & 82.8 & 71.3 \\
    \midrule
    \multirow{4}{*}{\rotatebox{90}{\textbf{Text}}} 
    & SBERT & FT & 82.0 & \textbf{81.6} & 42.4 & \textbf{95.2} & 45.6 & \textbf{94.4} &  &  &  &  & 44.0 & \textbf{94.8} \\
    & CLIP & FT & 88.4 & 75.2 & 42.8 & 94.0 & 64.0 & 91.2 &  &  &  &  & 53.4 & 92.6 \\
    & TBM & FT & 83.2 & 64.8 & \textbf{69.2} & 68.4 & \textbf{73.6} & 78.8 &  &  &  &  & \textbf{71.4} & 73.6 \\
    & \textbf{MiRAGe-T} & FT & \textbf{90.4} & 76.0 & 53.6 & 91.6 & 65.2 & 91.6 &  &  &  &  & 59.4 & 91.6 \\
    \midrule
    \multirow{5}{*}{\rotatebox{90}{\textbf{Image + Text}}} 
    & GPT-4o & ZS & \textbf{100.0} & 3.6 & \textbf{100.0} & 80.2 & \textbf{100.0} & 85.5 & \textbf{100.0} & 25.3 & \textbf{100.0} & 41.1 & \textbf{100.0} & 58.0 \\
    & Gemini 1.5 & ZS & 98.8 & 14.0 & 89.8 & 68.0 & 86.6 & 75.8 & 89.9 & 38.5 & 86.8 & 51.4 & 88.3 & 58.4 \\
    & \multirow{2}{*}{HAMMER} & ZS & 70.0 & 8.0 & 83.6 & 14.8 & 75.2 & 15.2 & 83.6 & 18.4 & 75.2 & 22.4 & 79.4 & 17.7 \\
    &  & FT & 99.6 & \textbf{99.2} & 92.8 & 61.6 & 94.8 & 62.8 & 92.8 & 77.2 & 94.8 & 81.6 & 93.8 & 70.8 \\
    & \textbf{MiRAGe} & FT & 98.4 & 98.4 & 80.8 & \textbf{90.0} & 72.0 & \textbf{87.2} & 80.8 & \textbf{91.2} & 72.0 & \textbf{95.6} & 76.4 & \textbf{91.0} \\
    \bottomrule
    \end{tabular}
    \caption{\textbf{Class-wise accuracy (\%) of all detectors across datasets.} "Real" and "Fake" columns represent the accuracy of the real and fake/generated classes, respectively. We leave SDXL columns blank in Text-only detectors since they share the same textual data as the DALL-E columns. We find that MLLMs are great at recognizing real images, especially with additional textual information. We also find that our model is much better at detecting generated image-caption pairs than other models}
    \label{tab:accuracy}
\end{table*}

\clearpage
\appendix

\section{Model Implementation Details}
\label{sec:appendix}

\subsection{Image-only Detector}
\textbf{Linear Model.} For our linear model, we use a frozen EVA-CLIP ViT encoder from BLIP to embed our images and add a linear layer with Sigmoid activation to project down to the output dimension.
\smallbreak
\noindent \textbf{Object Class CBM.} Concept Bottleneck Models (CBM) \cite{koh_concept_2020, yang2023language} have been shown to improve generalizability in image classification tasks. Extending this approach with the intuition that anomalies in generated images are often regional and object-based (i.e., merged fingers, curved buildings), we choose common object classes as the concepts. 

We first utilize OwlV2 to detect and crop out the objects in both real and fake images. These crops are organized into a dataset of 300 object classes, each containing real or fake crops of the object. We then train a linear model to detect fake objects within each object class. 

To create a list of concept scores for each image as a bottleneck, with each detected object class, we use the corresponding classifier to predict each instance of the object class and keep the maximum prediction score. With undetected object classes having a prediction score of 0, we map any input image to 300 concept scores. Lastly, we train a linear model as the bottleneck predictor and only use the 300 concept scores to predict a given image. 

\smallbreak
\noindent \textbf{MiRAGe-Img} ensembles the linear model and Object Class CBM. Our experiments on Object Class CBM show that, although having lower overall accuracy, it learns to detect fake images much better than the linear model (+14.2\% Fake Accuracy). Incorporating the strengths of both models, we compute the prediction score from the linear model as an extra concept score and concatenate it with the original 300 concept scores. This model achieves state-of-the-art performance in our testing set.

\begin{table}[t]
    \small
    \centering
    \begin{tabular}{p{2.7cm}|p{0.8cm}|p{2.95cm}}
    \toprule
    \textbf{Model}&\textbf{Task}&\textbf{Method Summary}\\
    % &\textbf{-L7B}&\textbf{-G2X}\\
    \midrule
    {\textbf{DE-FAKE}\newline\cite{defake}}&Image&Uses BLIP to caption images and uses extra text feats. to detect\\{\textbf{DIRE}\newline\cite{wang2023dire}}&Image&Computes image reconstruction error during diffusion and denoising\\
    {\textbf{KNN}\newline\cite{ojha23}}&Image&Maps real and fake img. to feat. space and uses the closest image to predict\\
    {\textbf{Linear (EVA-CLIP)}\newline\cite{evaclip}}&Image&Uses EVA-CLIP to encode images and trains a linear model\\
    \textbf{Obj-Class CBM}&Image&Trains one classifier per object class, and predicts w/ outputs\\
    \midrule
    {\textbf{Linear (SBERT)}\newline\cite{sentencebert}}&Text&Uses SentenceBERT to encode captions and trains a linear model\\ 
    {\textbf{Linear (CLIP)}\newline\cite{clip}}&Text&Uses CLIP to encode captions and trains a linear model\\
    {\textbf{TBM}\newline\cite{tbm}}&Text&Prompts LLM to discover textual concepts and trains a linear layer\\
    \midrule
    {\textbf{HAMMER}\newline\cite{shao2023detectinggroundingmultimodalmedia}}&Image + Text&Designs a Manipulation Aware Contrastive Learning Loss to capture the semantic inconsistency between images and text.\\
    \bottomrule
    \end{tabular}
    \caption{Summaries of detectors used in our image-only and text-only detection.}
    \label{tab:detectors}
\end{table}

\subsection{Text-only Detectors}

\textbf{Linear Model.} Similar to the linear model for images, we add a linear layer with sigmoid activation on top of a frozen pre-trained text encoder. We explored various text encoders and surprisingly found that the CLIP encoder performed better than Sentence BERT \cite{sentencebert}.

\noindent \textbf{TBM (Text Bottleneck Model).} harnesses the power of LLM to automatically extract distinguishing concept features from the text. We adopt this method and iteratively extract 18 diverse concepts from our captions after filtering. We then train a linear layer as a bottleneck predictor to map the concept scores to final predictions.

\noindent \textbf{MiRAGe-Txt} ensembles the linear model and Text CBM. Our experiments on Text CBM show a similar pattern: while it has lower overall accuracy, it is better at detecting real captions than the linear model (+10\% Fake ACC). We ensemble these two models by adding a prediction score from the linear model as an extra concept score to the bottleneck and training another linear layer on top for binary classification.

\subsection{Multimodal Detectors}
In this multimodal task, \textbf{MiRAGe} combines outputs from \textbf{MiRAGe-Img} and \textbf{MiRAGe-Txt} to predict if an image-caption pair is real or generated. We explored both early fusion and later fusion approaches:

\noindent \textbf{Early Fusion.} In this approach, we concatenate both image and text features from a given pair and train a linear model to predict. This method allows the model to use information from both modalities in conjunction to make predictions.

\noindent \textbf{Late Fusion.} In this approach, we first utilize the image-only and caption-only models to make corresponding predictions. We take the average of the logits from these two models with a sigmoid activation to make a single prediction of the pair.

\begin{table*}
    \small
    \centering
    \setlength{\tabcolsep}{2.3pt} % Adjusts the space between columns
    \begin{tabular}{c p{2.5cm} |ccc|ccc|ccc|ccc|ccc|ccc}
    \toprule 
    & &\multicolumn{3}{c|}{\textbf{NYT+MJ (ID)}}&\multicolumn{3}{c|}{\textbf{BBC+DALL-E}}&\multicolumn{3}{c|}{\textbf{CNN+DALL-E}}&\multicolumn{3}{c|}{\textbf{BBC+SDXL}}&\multicolumn{3}{c|}{\textbf{CNN+SDXL}}&\multicolumn{3}{c}{\textbf{OOD-AVG}}\\
    & &\textbf{Acc}&\textbf{F1}&\textbf{AP}&\textbf{Acc}&\textbf{F1}&\textbf{AP}&\textbf{Acc}&\textbf{F1}&\textbf{AP}&\textbf{Acc}&\textbf{F1}&\textbf{AP}&\textbf{Acc}&\textbf{F1}&\textbf{AP}&\textbf{Acc}&\textbf{F1}&\textbf{AP}\\
    \midrule
    \multirow{3}{*}{\rotatebox{90}{\textbf{Image}}} 
    
    & \textbf{MiRAGe-I} & \textbf{98.0} & \textbf{98.0} & \textbf{99.9} & \textbf{77.6} & \textbf{75.0} & \textbf{88.4} & \textbf{74.6} & 73.8 & 84.4 & \textbf{78.4} & \textbf{76.1} & \textbf{87.7} & \textbf{77.6} & \textbf{77.6} & \textbf{83.0} & \textbf{77.1} & \textbf{75.6} & \textbf{85.9} \\
    & (\textit{-CBM})& 97.6 & 97.6 & \textbf{99.9} & 70.4 & 63.0 & 84.2 & 66.6 & 62.3 & 75.1 & 77.2 & 73.7 & 87.3 & 74.2 & 73.2 & 81.2 & 72.1 & 68.1 & 82.0 \\
    & (\textit{-Linear}) & 94.2 & 94.2 & 98.1 & 66.6 & 70.3 & 82.8 & 72.0 & \textbf{74.4} & \textbf{84.6} & 65.6 & 69.2 & 74.9 & 65.0 & 65.8 & 73.8 & 67.3 & 69.9 & 79.0 \\
    \midrule
    
    \multirow{3}{*}{\rotatebox{90}{\textbf{Text}}}
    & \textbf{MiRAGe-T} & \textbf{83.2} & \textbf{81.9} & \textbf{91.0} & \textbf{72.6} & \textbf{77.0} & \textbf{85.2} & \textbf{78.4} & \textbf{80.9} & \textbf{92.0} & & & & & & & \textbf{75.5} & \textbf{79.0} & \textbf{88.6} \\
    & (\textit{-TBM}) & 81.8 & 80.5 & 90.2 & 68.4 & 74.8 & 84.4 & 77.6 & 80.3 & 90.9 & & & & & & & 73.0 & 77.6 & 87.7 \\
    & (\textit{-Linear}) & 74.0 & 71.4 & 78.1 & 68.8 & 68.7 & 72.1 & 76.2 & 76.8 & 75.2 & & & & & & & 72.5 & 72.8 & 73.7 \\
    
    \midrule
    
    \multirow{4}{*}{\rotatebox{90}{\textbf{Img+Txt}}} 
    & \textbf{MiRAGe} & 98.4 & 98.4 & \textbf{99.9} & \textbf{85.4} & \textbf{86.0} & \textbf{94.1} & \textbf{79.6} & \textbf{81.0} & \textbf{90.4} & \textbf{86.0} & \textbf{86.7} & \textbf{95.9} & 83.8 & 85.5 & \textbf{94.7} & \textbf{83.7} & \textbf{84.8} & \textbf{93.8} \\
    
    & (\textit{-Late Fusion}) & 98.4 & 98.4 & \textbf{99.9} & 80.6 & 78.4 & 92.7 & 79.4 & 78.0 & 89.3 & 83.2 & 81.8 & 93.9 & \textbf{85.8} & \textbf{85.8} & 93.4 & 82.3 & 81.0 & 92.3 \\
    
    & (\textit{-CBMs}) & \textbf{99.0} & \textbf{99.0} & \textbf{99.9} & 81.4 & 81.9 & 90.6 & 72.4 & 73.8 & 82.5 & 85.0 & 85.9 & 94.7 & 80.8 & 83.1 & 91.8 & 79.9 & 81.2 & 89.9 \\
    & (\textit{-Linears}) & 94.6 & 94.6 & 98.8 & 71.0 & 74.4 & 84.4 & 74.0 & 76.8 & 84.8 & 69.8 & 73.1 & 77.5 & 75.2 & 78.1 & 79.5 & 72.5 & 75.6 & 81.6 \\
    
    \bottomrule
    \end{tabular}
    \caption{\textbf{The Accuracy, F-1 score, and Average Precision (AP) for ablation study on MiRAGe-Img, MiRAGe-Txt, and MiRAGe.} We find that while all components of the MiRAGe model are central to its high performance, the linear models perform better than the concept bottleneck models, and early fusion performs slightly worse than late fusion.}
    \label{tab:ablation-res}
\end{table*}

\section{Ablation Study}
\label{sec:ablation}

As shown in Table \ref{tab:ablation-res}, all components of the MiRAGe detector are essential for their performance. We find that the linear models perform better than the concept bottleneck models, and early fusion performs slightly worse than late fusion.

We further investigate the class-wise accuracy for each component as shown in Table \ref{tab:albation-accuracy}. We find that the linear model is better at recognizing real images, while Obj-CBM is better at detecting fake images. We also see that the linear model is better at detecting fake captions, while TBM is better at recognizing real captions. This finding gives us some idea as to why the ensemble methods (MiRAGe-Img and MiRAGe-Txt) perform better than their individual components.

While CBMs always underperform linear models, they help them in the lower-performant class without affecting the higher-performant class when ensembling in our MiRAGe detector. However, the tradeoff is that ensembling an interpretable method with a black-box method takes away the interpretability.

\begin{table*}[b]
    \small
    \centering
    \setlength{\tabcolsep}{2.3pt} % Adjusts the space between columns
    \begin{tabular}{c p{2.5cm} |cc|cc|cc|cc|cc|cc}
    \toprule 
     & &\multicolumn{2}{c|}{\textbf{NYT+MJ (ID)}}&\multicolumn{2}{c|}{\textbf{BBC+DALL-E}}&\multicolumn{2}{c|}{\textbf{CNN+DALL-E}}&\multicolumn{2}{c|}{\textbf{BBC+SDXL}}&\multicolumn{2}{c|}{\textbf{CNN+SDXL}}&\multicolumn{2}{c}{\textbf{OOD-AVG}}\\
     & &\textbf{Real}&\textbf{Fake}&\textbf{Real}&\textbf{Fake}&\textbf{Real}&\textbf{Fake}&\textbf{Real}&\textbf{Fake}&\textbf{Real}&\textbf{Fake}&\textbf{Real}&\textbf{Fake}\\
    \midrule
    \multirow{3}{*}{\rotatebox{90}{\textbf{Image}}}

    & \textbf{MiRAGe-I} & \textbf{98.8} & \textbf{97.2} & 88.0 & 67.2 & 77.6 & 71.6 & 88.0 & 68.8 & 77.6 & \textbf{77.6} & 82.8 & 71.3 \\
    & (\textit{-CBM})  & 98.8 & 96.4 & \textbf{90.4} & 50.4 & \textbf{78.0} & 55.2 & \textbf{90.4} & 64.0 & \textbf{78.0} & 70.4 & \textbf{84.2} & 60.0 \\
    & (\textit{-Linear}) & 94.8 & 93.6 & 54.0 & \textbf{79.2} & 62.8 & \textbf{81.2} & 54.0 & \textbf{77.2} & 62.8 & 67.2 & 58.4 & \textbf{76.2} \\
    \midrule
    \multirow{3}{*}{\rotatebox{90}{\textbf{Text}}} 
    & \textbf{MiRAGe-T} & \textbf{90.4} & \textbf{76.0} & 53.6 & 91.6 & 65.2 & \textbf{91.6} &  &  &  &  & 59.4 & 91.6 \\
    & (\textit{-TBM})  & 88.4 & 75.2 & 42.8 & \textbf{94.0} & 64.0 & 91.2 &  &  &  &  & 53.4 & \textbf{92.6} \\
    & (\textit{-Linear})& 83.2 & 64.8 & \textbf{69.2} & 68.4 & \textbf{73.6} & 78.8 &  &  &  &  & \textbf{71.4} & 73.6 \\
    \midrule
    \multirow{4}{*}{\rotatebox{90}{\textbf{Img+Txt}}} 
    
    & \textbf{MiRAGe}  & 98.4 & 98.4 & 80.8 & \textbf{90.0} & 72.0 & \textbf{87.2} & 80.8 & \textbf{91.2} & 72.0 & \textbf{95.6} & 76.4 & \textbf{91.0} \\
    & (\textit{-Late Fusion}) & \textbf{99.6} & 97.2 & \textbf{90.8} & 70.4 & \textbf{85.6} & 73.2 & \textbf{90.8} & 75.6 & \textbf{85.6} & 86.0 & \textbf{88.2} & 76.3 \\
    & (\textit{-CBMs})  & 98.8 & \textbf{99.2} & 78.4 & 84.4 & 67.2 & 77.6 & 78.4 & 91.6 & 67.2 & 94.4 & 72.8 & 87.0 \\
    & (\textit{-Linears})  & 95.2 & 94.0 & 57.6 & 84.4 & 62.0 & 86.0 & 57.6 & 82.0 & 62.0 & 88.4 & 59.8 & 85.2 \\
    
    \bottomrule
    \end{tabular}
    \caption{\textbf{Class-wised accuracy for ablation study on MiRAGe-Img, MiRAGe-Txt, and MiRAGe.}  "Real" and "Fake" columns represent the accuracy of the real and fake/generated classes, respectively. EF and LF stand for Early Fusion and Late Fusion, respectively. We leave SDXL columns blank in Text-only detectors since they share the same textual data as the DALL-E columns. We find that linear models and CBMs are good at classifying different classes. We also find that different fusion techniques lead to different strengths of the models}
    \label{tab:albation-accuracy}
\end{table*}

\section{Human Study Details}

\begin{figure}
    \centering
    \includegraphics[width=\columnwidth]{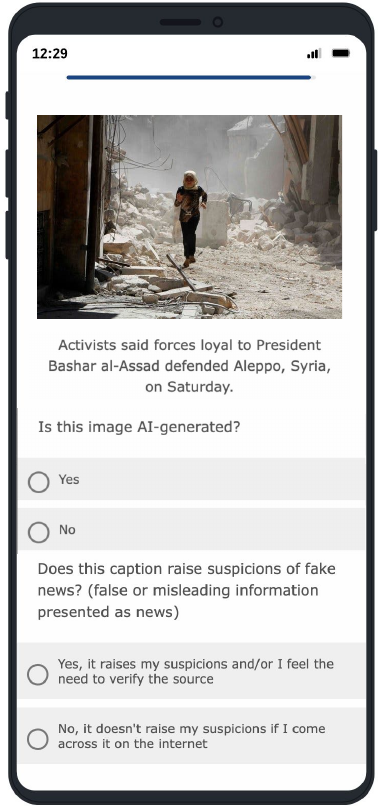}
    \caption{The user interface of the human study where each participant is given a pair of news images and caption and asked to determine whether they are real or generated}
    \label{fig:ui}
\end{figure}

We recruited 112 graduate students enrolled in a CS class with extra credits as compensation, and each participant was randomly assigned 20 image-caption pairs to annotate. They are asked to determine whether the image is generated and whether the caption raises their suspicions of fake news, as shown in Figure \ref{fig:ui}. Each image-caption pair in the survey dataset is shown to three participants, and the majority of the votes determine the human's judgment. Note that if a participant thinks either the image or the caption is generated, the final decision of the news would be ``generated''.

The results show an overall accuracy of 71.4\%, F-1 score of 60.4\% and precision of 89\%. Moreover, humans are much better at recognizing real news (97\% real acc.) than detecting generated news (45.8\% fake acc.). The Krippendorff's Alpha from the annotations is 0.22, which shows a low agreement among the participants and implies that it is difficult for humans to consistently detect generated news. These results further reflects the danger of such hyperrealistic generated news.

\begin{figure}[t]
  \begin{minipage}{\columnwidth} % Start the first minipage
    \centering
    \includegraphics[width=\columnwidth]{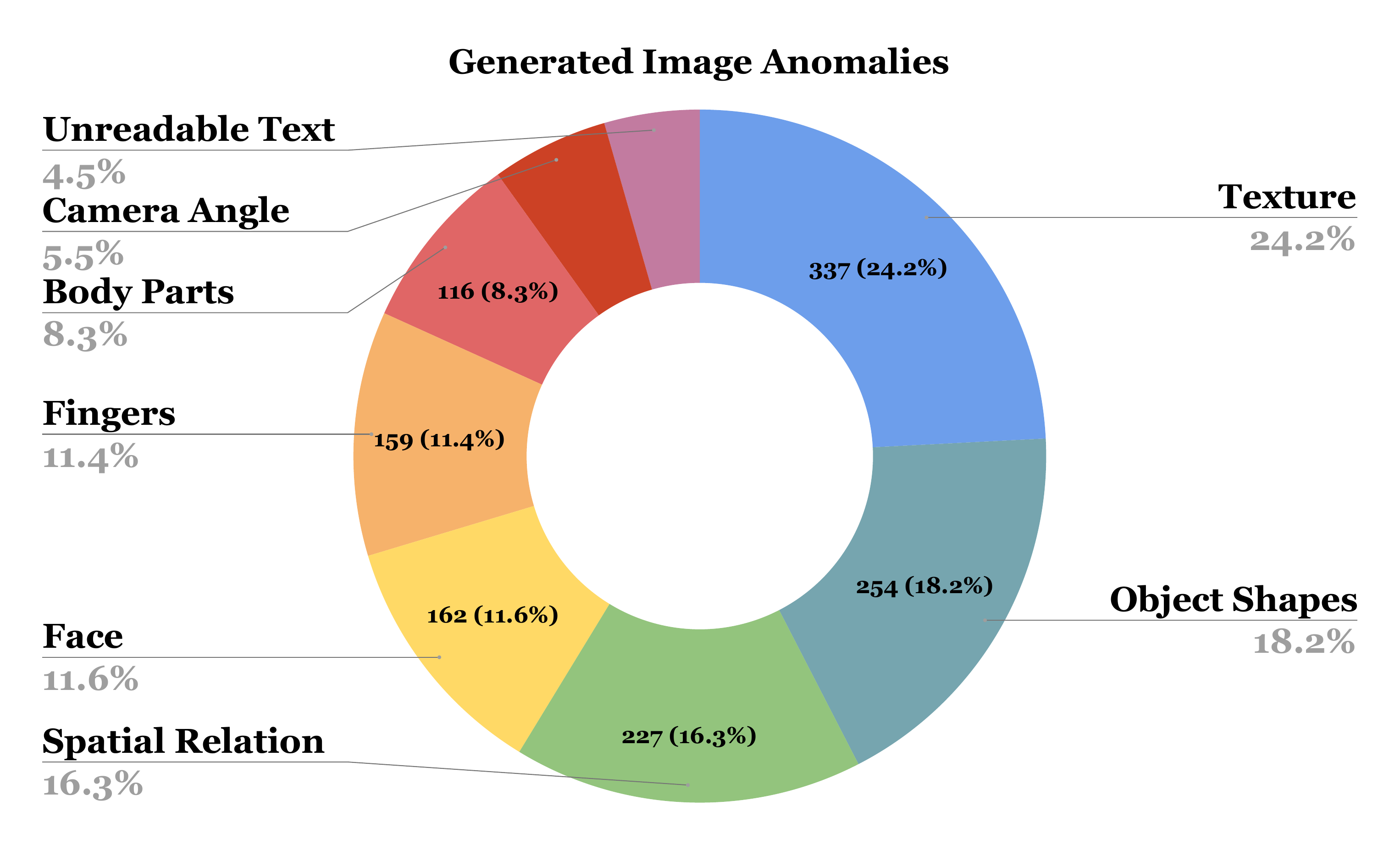}
    \caption{Annotated generated image anomalies from human study}
    \label{fig:img-reason}
  \end{minipage}

  \vspace{1em} % Adjust the vertical space between the two figures

  \begin{minipage}{\columnwidth} % Start the second minipage
    \centering
    \includegraphics[width=\columnwidth]{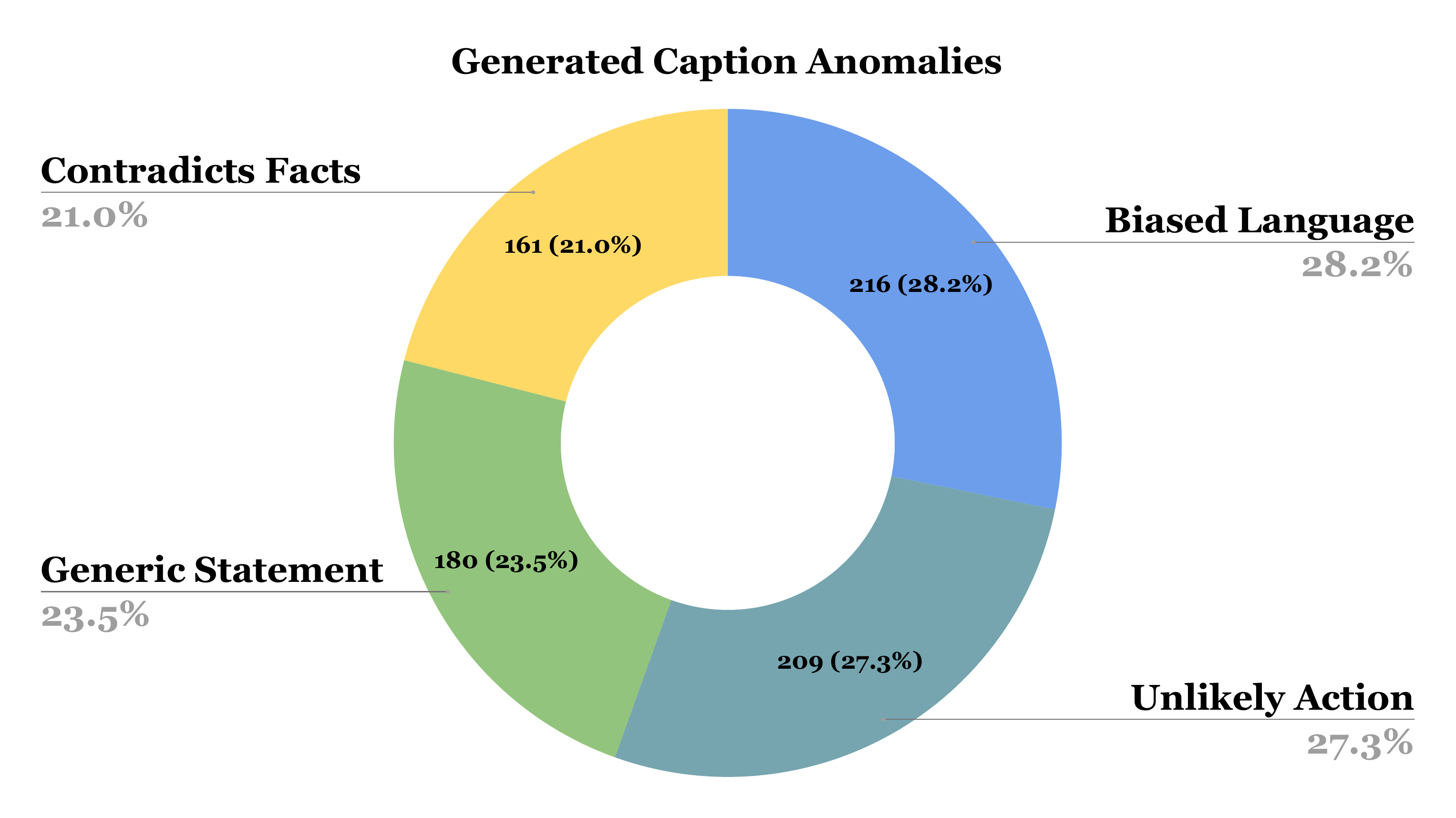}
    \caption{Annotated generated caption anomalies from human study}
    \label{fig:cap-reason}
  \end{minipage}
\end{figure}

The participants are also asked to provide the reasons why they think a given image or caption is generated. We provide a list of common anomalies found in generated images and captions to choose from as shown in Table \ref{tab:anomalies}. For the generated images that the participants correctly classified, the top clues humans are using are Texture (24.2\%), Object Shapes (18.2\%), and Spatial Relation (16.3\%) as shown in Figure \ref{fig:img-reason}. Similarly, the top clues for generated captions are Biased Language (28.2\%) and Unlikely Action (27.3\%) as shown in Figure \ref{fig:cap-reason}.

\begin{table}[t]
    \small
    \centering
    \begin{tabular}{p{0.1cm} p{2.3cm}|p{3.95cm}}
    \toprule
    &\textbf{Anomaly}&\textbf{Detail Description}\\
    % &\textbf{-L7B}&\textbf{-G2X}\\
    \midrule
    \multirow{15}{*}{\rotatebox{90}{\textbf{Image}}}
    &Texture&Unrealistically perfect and smooth texture\\
    &Object Shapes&Irregular shape or composition of objects\\
    &Spatial Relation&Illogical or impossible spatial relation of objects or people\\
    &Face&Unnatural facial features or expressions\\
    &Fingers&Incorrect number of fingers or twisted or merged fingers\\
    &Body Parts&Extra or missing body parts or merged or deformed body parts\\
    &Camera Angle&Irregular or impossible camera angle\\
    &Unreadable Text&Unreadable or misspelled text\\
    \midrule
    \multirow{9}{*}{\rotatebox{90}{\textbf{Caption}}}
    &Biased Language&Uses suspiciously biased or exaggerated language\\
    &Unlikely Action&Has action that is unlikely or impossible to be performed by the subject\\
    &Generic Statement&Uses generic statement that lacks necessary details\\
    &Contradicts Facts&Contradicts with known facts or events\\
    \bottomrule
    \end{tabular}
    \caption{Anomalies choices provided in the human study. Participants can choose more than one reason.}
    \label{tab:anomalies}
\end{table}

\section{Examples}
\begin{figure*}[t]
    \centering
    \includegraphics[width=\textwidth]{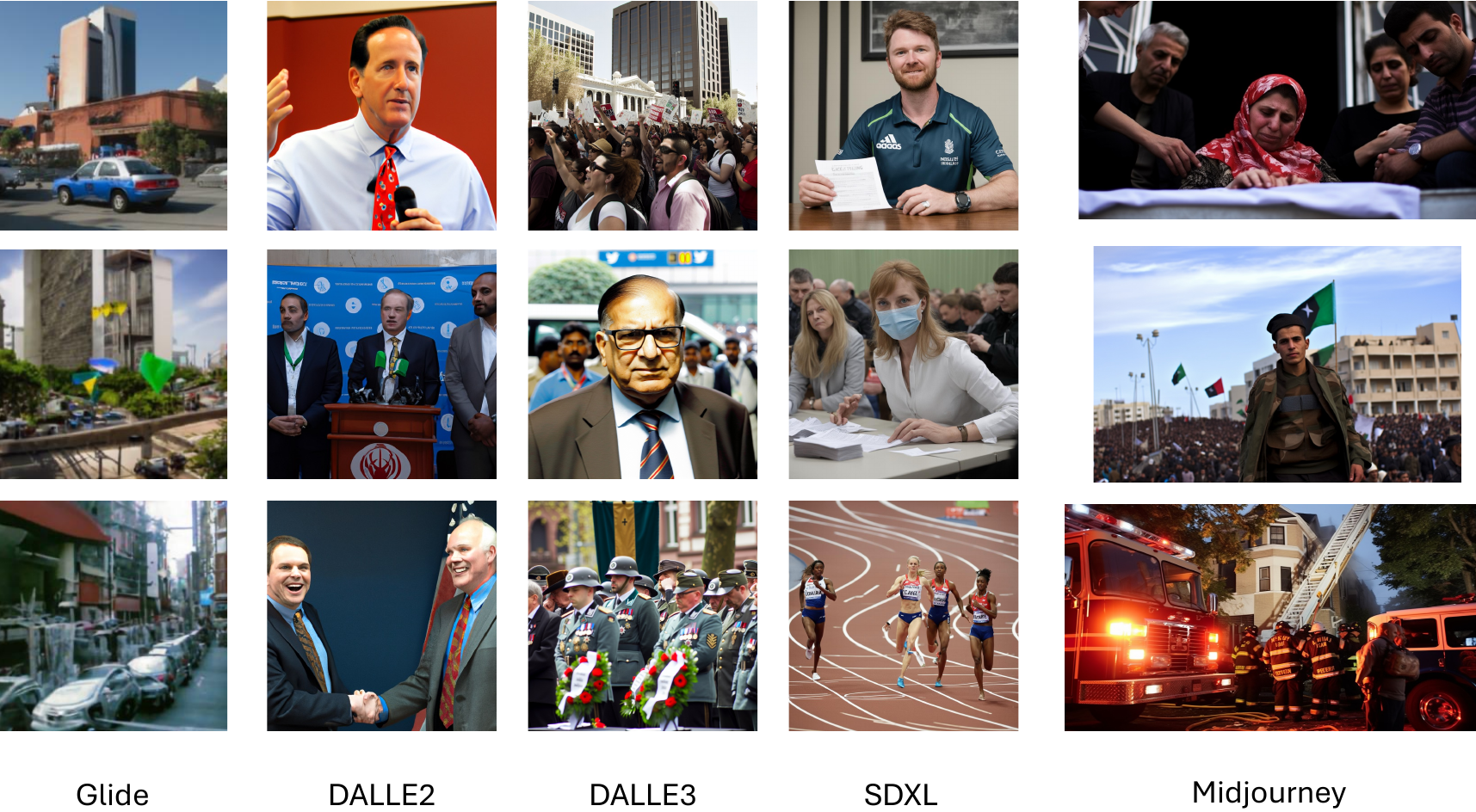}
    \caption{Comparison of different image generators across examples from the MiRAGeNews dataset. We see that modern generators such as Midjourney produce high-quality images that are difficult to distinguish from real images.}
    \label{fig:pics}
\end{figure*}

In Figure \ref{fig:pics} we show examples from the MiRAGeNews dataset from different image generators. We see that Midjourney produces images that are much more difficult to detect than other similar generators. Combined with the lack of moderation, we feel that this generator is the most important to cover in a dataset such as ours.

\end{document}